\documentclass[a4paper,12pt]{article}
\usepackage{tabularx}
\usepackage{booktabs} 
\usepackage{caption} 
\usepackage{graphicx} 

\usepackage[left=2cm,right=2cm,top=2cm,bottom=2cm]{geometry}
\usepackage[utf8]{inputenc}

\usepackage{setspace}
\usepackage{mathptmx} 

\usepackage[hidelinks]{hyperref} 
\usepackage{apacite} 

\usepackage{geometry}
\usepackage{multirow}
\usepackage{amsmath}
\usepackage{float} 
\usepackage{siunitx}

\usepackage{tabularx} 
\usepackage{color,soul}

\usepackage{orcidlink}

\title{{\tiny Author's Manuscript. Final Version of Record published in Digital Management and Artificial Intelligence. Proceedings of the Fourth International Scientific-Practical Conference (ISPC 2024), Hybrid, October 10-11, 2024, available online at:  \href{https://link.springer.com/chapter/10.1007/978-3-031-88052-0_45}{\url{https://link.springer.com/chapter/10.1007/978-3-031-88052-0_45}}} \newline \newline \large{Cyber Security Data Science: Machine Learning Methods and their Performance on Imbalanced Datasets}}

\author{ Mateo Lopez-Ledezma\orcidlink{0009-0001-0268-4114}\thanks{
Computational Systems Engineering, Universidad Privada Boliviana, Bolivia. E-mail: lopezmateo97@yahoo.com} $ $ and Gissel Velarde\orcidlink{0000-0001-5392-9540}\thanks{Extended Artificial Intelligence, IU International University of Applied Sciences, Erfurt, 99084, Germany E-mail: gissel.velarde@iu.org} }  
\date{}

\begin{document}

\maketitle

\begin{abstract}
  \small  Cybersecurity has become essential worldwide and at all levels, concerning individuals, institutions, and governments. A basic principle in cybersecurity is to be always alert. Therefore, automation is imperative in processes where the volume of daily operations is large. Several cybersecurity applications can be addressed as binary classification problems, including anomaly detection, fraud detection, intrusion detection, spam detection, or malware detection. In many cases, the positive class samples, those that represent a problem, occur at a much lower frequency than negative samples, and this poses a challenge for machine learning algorithms since learning patterns out of under-represented samples is hard. This is known in machine learning as imbalance learning.  In this study, we systematically evaluate various machine learning methods using two representative financial datasets containing numerical and categorical features. The Credit Card dataset contains $283\,726$ samples, 31 features, and 0.2 percent of the transactions are fraudulent  (imbalance ratio of 598.84:1). The PaySim dataset contains $6\,362\,620$ samples, 11 features and 0.13 percent of the transactions are fraudulent (imbalance ratio of 773.70:1). We present three experiments. In the first experiment, we evaluate single classifiers including Random Forests, Light Gradient Boosting Machine, eXtreme Gradient Boosting, Logistic Regression, Decision Tree, and Gradient Boosting Decision Tree. In the second experiment, we test different sampling techniques including over-sampling, under-sampling, Synthetic Minority Over-sampling Technique, and Self-Paced Ensembling. In the last experiment, we evaluate Self-Paced Ensembling and its number of base classifiers. We found that imbalance learning techniques had positive and negative effects, as reported in related studies. Thus, these techniques should be applied with caution. Besides, we found different best performers for each dataset. Therefore, we recommend testing single classifiers and imbalance learning techniques for each new dataset and application involving imbalanced datasets as is the case in several cyber security applications. We provide the code with all experiments as open-source.\footnote{Available at \url{https://github.com/MateoLopez00/Imbalanced-Learning-Empirical-Evaluation}.}
\end{abstract}


\section{Introduction}
Cyber security is gaining prominence worldwide as cybercrime increases. The cost of cybercrime to the global economy is estimated at 400 billion USD~\cite{sarker2020cybersecurity}. Cyber security data science aims at discovering patterns from data to detect anomalous, malicious, or fraudulent events that can harm a system. Applications in cyber security include anomaly detection, fraud detection, intrusion detection, spam detection, malware detection, and phishing detection~\cite{sarker2020cybersecurity, shaukat2020performance}. Cyber security systems should detect possible attacks at any time, even if the number of operations increases suddenly. Availability and scalability pose a challenge to systems that depend on human monitoring, therefore, automation in cyber security is wanted. Artificial Intelligence (AI), and more specifically, machine learning allows intelligent automation based on the available data to detect cybercrime activities automatically. 

Cyber security is now a trending topic just like data science and machine learning~\cite{sarker2020cybersecurity}. In some countries, the term data science is more popular than machine learning and vice versa. Some authors consider that data science would encompass machine learning~\cite{sarker2020cybersecurity}. \citeA{sculley2015hidden} argue that the code devoted to purely machine learning algorithms in real-world systems represents a small percentage of the required infrastructure consisting of several tasks including configurations, data collection, feature extraction, data verification, analysis tools, process and machine resource management tools, serving infrastructure and monitoring.   Still, the distinction between what is considered data science and machine learning is often not clear, and these terms may be used as synonyms in some fields or as complementary terms in other domains \cite{alpaydin2014introduction}. What is certain is that cyber security is a must for individuals, institutions, and governments, as cyber-attacks increase.   Therefore, extracting patterns from data to automatically detect attacks can only be done computationally to stay alert  24 hours a day, 7 days a week. However, in addition to availability and scalability, a fundamental component is the effectiveness of a system at detecting attacks. Therefore, detection quality depends on the performance of the machine learning methods in place.

In this paper, we present machine learning methods and their performance on two representative imbalanced datasets, as many cyber security applications will face the problem of learning from imbalanced distributions. Several cyber security machine learning applications can be addressed as binary classification problems, where the positive class samples represent a problem to a system. In many cases, the positive samples are under-represented in the datasets, and therefore, it is hard to learn statistics from the minority class \cite{JMLR:v18:16-365, kim2022, hajek2022fraud, li2023imbalanced, VELARDE2024200354, yang2021progressive}. 

We summarise the contributions of this study as follows:
\begin{itemize}
\item We systematically evaluate the performance of six machine learning algorithms in terms of recognition and speed in two representative cyber security datasets for fraud detection. 
\item We test the effects of imbalance learning techniques including sampling and ensembling.
\item We show that imbalance classification depends strongly on the characteristics of each dataset, such that different classifiers should be evaluated for each dataset to select the best approach. 
\end{itemize}

\section{The Method}
\begin{figure*}[]
    \centering
    \includegraphics[width=18cm]{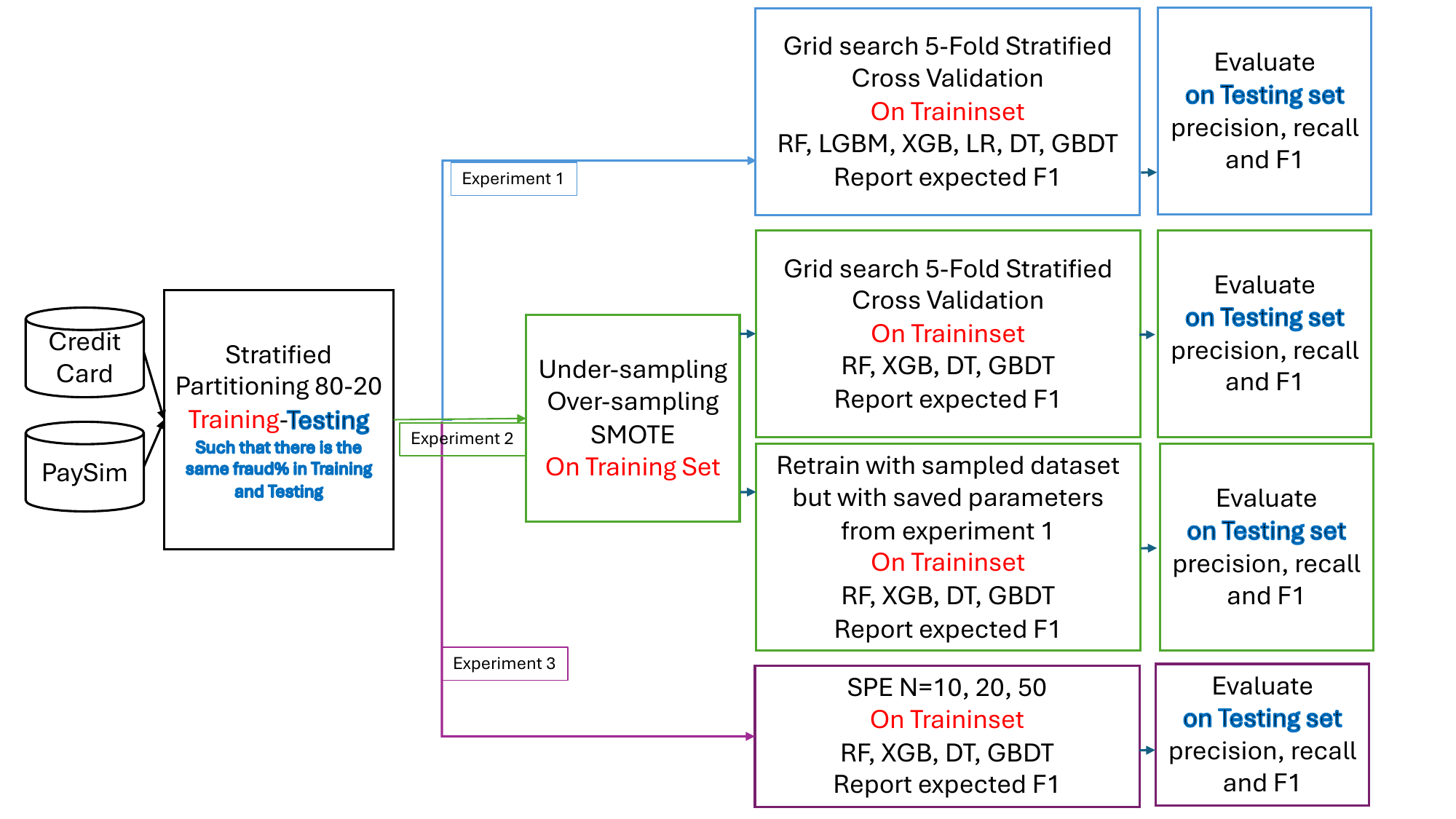}
    \caption{A schematic representation of the method.}.
    \label{f:experiments}
    
\end{figure*}

The proposed method can be seen in Figure \ref{f:experiments}. Each dataset follows a stratified partition of 80-20 percent for training-testing sets, respectively. In the first experiment, we evaluate the performance of single classifiers: Random Forests (RF), Light Gradient Boosting Machine (LGBM), eXtreme Gradient Boosting (XGB), Logistic Regression (LR), Decision Tree (DT), and Gradient Boosting Decision Tree (GBT), see section \ref{s:classifiers} for details. Each classifier goes under grid search on 5-Fold Stratified cross-validation on the training set, optimizing F1 score. In the second experiment, we evaluate the effect of sampling the training set to balance it, such that it has the same number of positive and negative samples, see section \ref{s:sampling}. First, we use the parameters found in Experiment 1 on the resampled training set. Then, we retrain the classifiers with the resampled training set in 5-fold Stratified cross-validation. In the third experiment, we evaluate Self-Paced Ensembling (SPE) with N=10, 20, and 50 base classifiers, see section \ref{s:sampling}.

Both datasets were analyzed and cleaned before classifier initialization. We performed exploratory data analysis to understand the datasets' structure, number of features, types of features, and imbalance ratio. The Credit Card dataset presented in total $1\,081$ duplicates but no missing (null) values.  To address duplicated values, the first occurrences of each duplicated row have been kept and subsequent occurrences have been removed for each dataset. PaySim dataset did not present missing (null) values or duplicated rows. 

\subsection{The datasets} \label{s:datasets}
We use two imbalanced datasets: \textbf{Credit Card} introduced in \cite{dal2017credit}, and \textbf{PaySim} introduced in \cite{lopez2016PaySim}, see Table \ref{tab:1}. Credit Card consists of $283\,726$ credit card transactions of a European e-commerce, where daily fraud transactions vary over time, but roughly account for 0.2 percent of the transactions \cite{dal2017credit}. PaySim consists of $6\,362\,620$ simulated transactions from a real dataset of African mobile payments. Fraud transactions account for 0.13 percent of all transactions \cite{lopez2016PaySim}.

\begin{table*}[htbp]
    \centering 
    \footnotesize
    \caption{Characteristics of Credit Card and PaySim Datasets.}
    \label{tab:1} 
        \begin{tabular}{@{}lcccccc@{}}
            \toprule
            \textbf{Dataset} & \textbf{\#Attributes} & \textbf{\#Samples} & \textbf{Feature Format} & \textbf{Imbalance Ratio} & \textbf{Fraud \%} \\
            \midrule
            Credit Card \cite{dal2017credit} & 31 & $283\,726$ & Numerical & 598.84:1  & 0.2 \\
            PaySim \cite{lopez2016PaySim} & 11 & $6\,362\,620$ & Numerical \& categorical & 773.70:1 & 0.13 \\
            \bottomrule
        \end{tabular}
\end{table*}

\subsection{Classifiers} \label{s:classifiers}
We evaluated the following classifiers:

\begin{itemize}

\item \textbf{Random Forests (RF)} are combinations of trees that depend on an independently sampled vector with the same distribution of the trees in a forest \cite{breiman2001}.

\item \textbf{Gradient Boosting Decision Tree (GBDT)} uses the principle of boosting to create a series of predictive models, each correcting the errors of its predecessors \cite{friedman2001greedy}. 

\item \textbf{Light Gradient Boosting Machine (LGBM)} aims at speeding up the training process of Gradient Boosting Decision Tree algorithms, delivering a scalable implementation by reducing the number of features \cite{ke2017lightgbm}. 

\item \textbf{eXtreme Gradient Boosting (XGB)} also belongs to the Gradient Boosting Decision Tree family, where successive models are added to correct the residuals or errors of previous models. This iterative approach minimizes the overall predictive error through a series of decision trees that progressively improve the model's accuracy \cite{chen2016xgboost}.

\item \textbf{Logistic Regression (LR)} is a statistical method for binary classification that estimates the probabilities of the outcomes based on a weight vector \cite{yu2011dual}. 

\item \textbf{Decision Tree (DT)} implements the divide-and-conquer strategy on a hierarchical data structure, and can be represented as a set of simple rules easy to interpret and visualize \cite{alpaydin2014introduction, quinlan1986induction}.
\end{itemize}

\subsection{Evaluation criteria} \label{s:metrics}

In binary imbalance classification, the minority class has a substantially less number of samples than the majority class. Identifying the minority class is hard but essential \cite{VELARDE2024200354}. Therefore, the minority class represents the positive class, and the majority class, the negative class.

\begin{table*}[]
\caption{Confusion matrix for binary classification.}
\label{tab:2}
\centering
\footnotesize 
    \begin{tabular}{l|l|c|c|c}
        \multicolumn{2}{c}{}&\multicolumn{2}{c}{Predicted}&\\
        \cline{3-4}
        \multicolumn{2}{c|}{}&Negative&Positive\\
        \cline{2-4}
        \multirow{2}{*}{Actual}& Negative & $TN$ & $FP$\\
        \cline{2-4}
        & Positive & $FN$ & $TP$\\
        \cline{2-4}
    \end{tabular}
\end{table*}

 Table \ref{tab:2} shows the confusion matrix for binary classification. The confusion matrix allows us to compute different evaluation metrics.  We do not consider accuracy as an evaluation metric, since it does not reflect well the model performance and may be deceiving for imbalanced scenarios \cite{saito2015precision, liu2020self, VELARDE2024200354}. The evaluation metrics considered in this study include  Recall = $\displaystyle\frac{TP}{TP + FN}$, Precision=$\displaystyle\frac{TP}{TP + FP}$, and F1 score=$\displaystyle 2 \cdot \frac{\text{Recall} \times \text{Precision}}{\text{Recall} + \text{Precision}}$. 

\subsection{Imbalance Learning techniques} \label{s:sampling}
We use four different imbalance learning techniques, including sampling and ensembling. Sampling techniques are used to change the training set original distribution.
\begin{itemize}

\item \textbf{Over-Sampling}: Aims at increasing the number of samples in the minority class by randomly resampling the minority class samples. This is a data augmentation technique. The size of the final training set after oversampling is two times the size of the original majority class, since we upsample the minority class until it is equal to the majority class. For example, for PaySim, the training set had $5\,083\,526$ majority samples, and $6\,570$ minority samples, we upsample the minority class so it also has $5\,083\,526$ samples. The size of the final training set is two times the size of the original majority class. 

\item \textbf{Synthetic Minority Over-sampling Technique (SMOTE)}:  Aims at balancing the class distribution by creating synthetic minority class examples and can under-sampling the majority class \cite{chawla2002smote}. In our case, the minority class was increased to reach the size of the majority class.  

\item \textbf{Under-Sampling}: Randomly removes samples from the majority class. The size of the final training set after under-sampling is two times the size of the original minority class, since we under-sample the majority class until it is equal to the minority class. For example for PaySim, the training set had $5\,083\,526$ majority samples, and 6570 minority samples, we under-sample the majority class so it also has $6\,570$ samples. The size of the final training set is two times the size of the original minority class. 



\item \textbf{Self-Paced Ensemble (SPE)} is an ensemble learning technique designed to handle imbalanced datasets by iteratively re-weighting the training data to focus on ``hard'' examples, aiming to improve recognition of the minority class \cite{liu2020self}. 
\end{itemize}
\section{Results}
The results of the three experiments are presented in this section. Both datasets, Credit Card and PaySim, are used independently in each experiment. See Appendix \ref{theappendix} for more information.

\subsection{Experiment 1. Individual classifiers.}
\begin{figure}[H]
    \centering
    \begin{minipage}{\textwidth}
        \centering
        \includegraphics[width=1\textwidth]{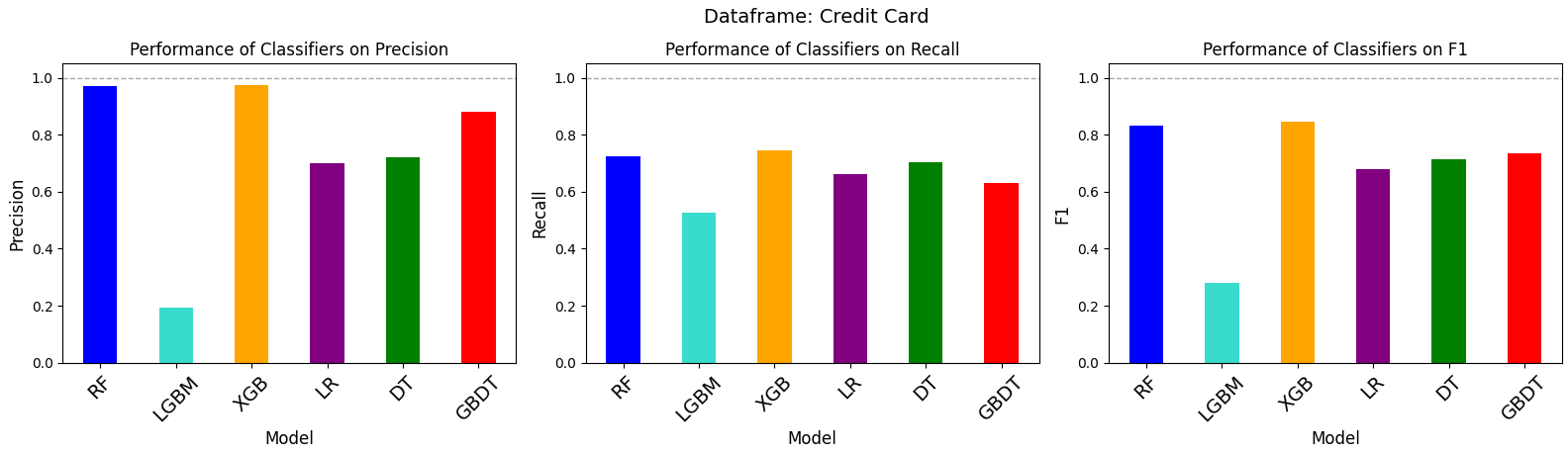} 
    \end{minipage}
    
    \vspace{1em} 
    
    \begin{minipage}{\textwidth}
        \centering
        \includegraphics[width=1\textwidth]{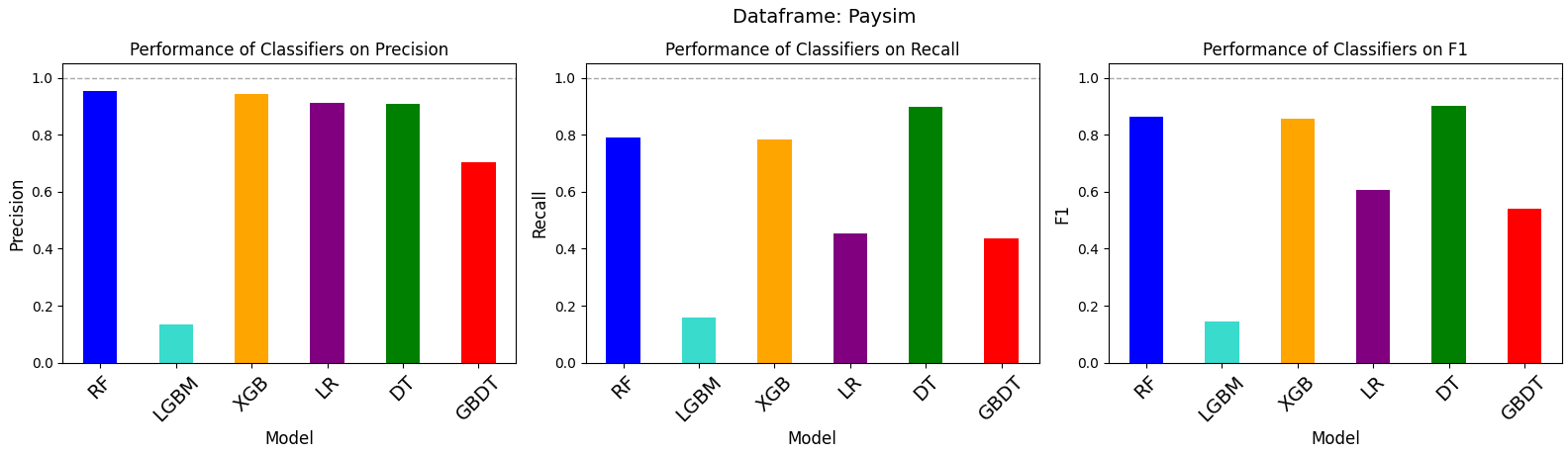} 
    \end{minipage}
    
    \caption{Experiment 1. Performance of individual classifiers evaluated over Precision, Recall, and F1. The upper row of figures corresponds to the results on Credit Card. The lower row corresponds to the results on PaySim.}
    \label{fig:fig1}
\end{figure}

Experiment 1 aims to evaluate individual classifiers without any sampling techniques. 
Figure \ref{fig:fig1} presents the experimental results of the six selected machine learning classifiers. XBG and RF are the best-performing classifiers, and their F1 score is similar in both datasets. However, XGB is much faster to train than RF, see Table \ref{tab:5}. GBDT and DT perform on a similar level on the Credit Card dataset, but DT outperforms all classifiers on PaySim. GDBT in contrast does a poor job on PaySim. LGBM is the weakest of all in both datasets. For the following experiments, we select four classifiers: XGB, RF, DT, and GBDT.


\subsection{Experiment 2. Imbalance Learning Methods}
\begin{figure}[H]
    \centering
    \begin{minipage}{\textwidth}
        \centering
        \includegraphics[width=1\textwidth]{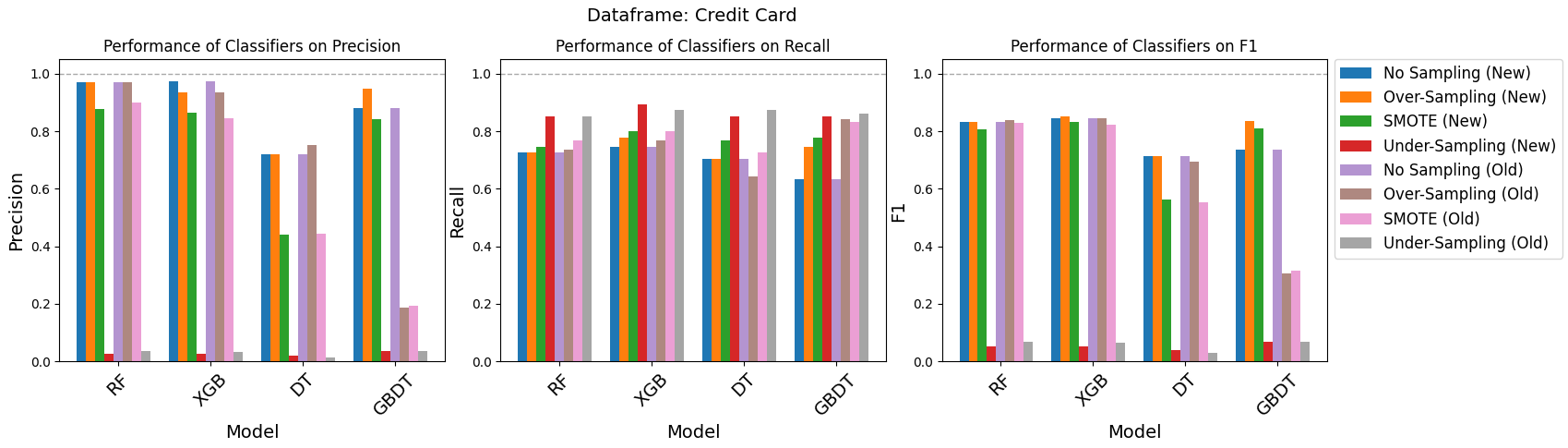} 
    \end{minipage}
    
    \vspace{1em} 
    
    \begin{minipage}{\textwidth}
        \centering
        \includegraphics[width=1\textwidth]{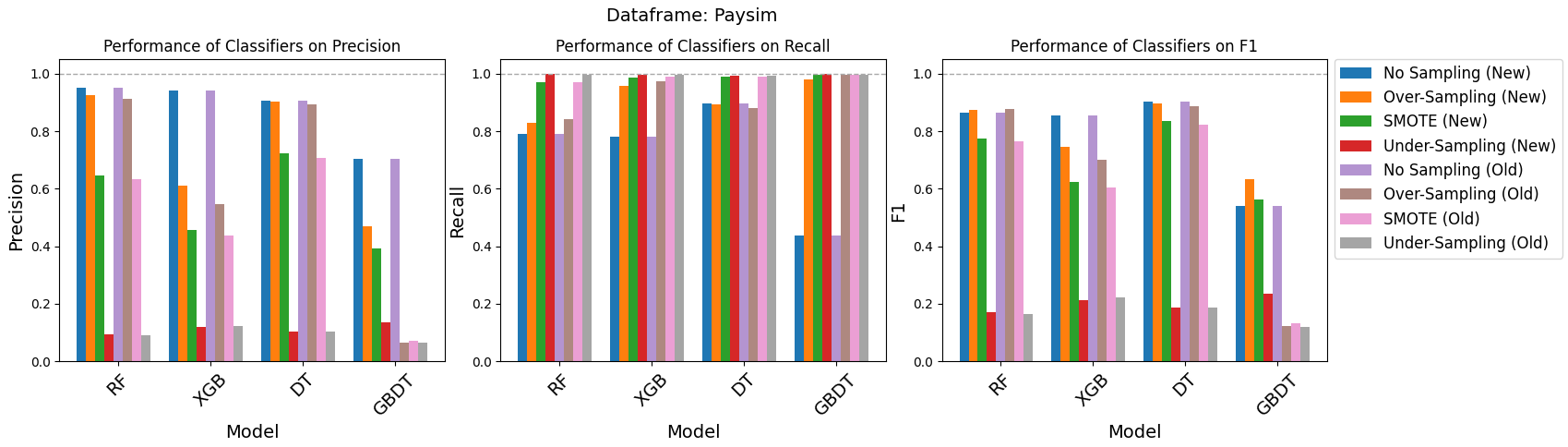} 
    \end{minipage}
    
    \caption{Experiment 2. Effect of sampling techniques: No Sampling, Over-Sampling, Under-Sampling, and SMOTE. Evaluation over Precision, Recall, and F1. Blue, orange, green, and red, marked as (New), correspond to optimising classifiers using the resampled training set. Purple, brown, pink, and grey, marked as (Old), correspond to using the parameters found in Experiment 1 and retraining with the resampled dataset. The upper row of figures corresponds to the results on Credit Card. The lower row corresponds to the results on PaySim.}
    \label{fig:fig2}
\end{figure}

Experiment 2 aims at evaluating the effect of sampling techniques, 
see Figure \ref{fig:fig2}. Compare the blue, orange, green, and red bars marked as (New) against the purple, brown, pink, and grey, marked as (Old) for No Sampling, Over-Sampling, SMOTE, and Under-Sampling, respectively. Sampling techniques have different effects. Over-Sampling has either no effect or helps improve recognition. SMOTE most of the time worsens F1, except for GBDT, such that it improves recognition only when optimization is done on the resampled dataset. More generally, sampling techniques improve recall at the expense of lowering precision. However, RF, XGB, and DT prove robust to resampling the training set. In contrast, GBDT performs better when optimising over the resampled dataset, such that the orange, green, and red bars marked as (New) achieve higher F1 score than the brown, pink, and grey, marked as (Old).

\subsection{Experiment 3. SPE and the number of base classifiers}

\begin{figure}[]
    \centering
    \begin{minipage}{\textwidth}
        \centering
        \includegraphics[width=1\textwidth]{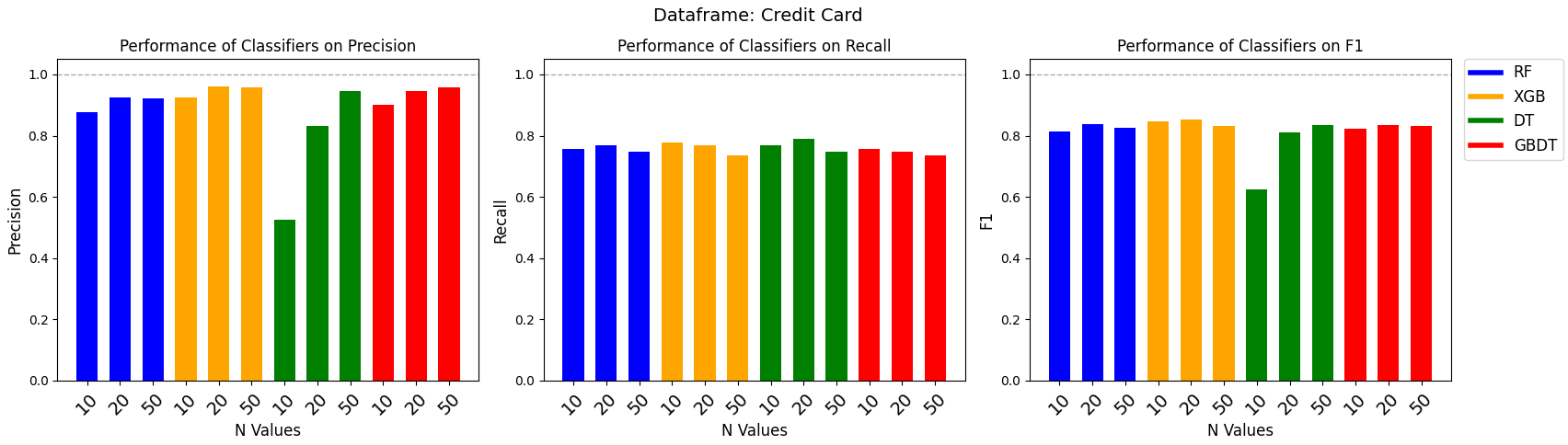} 
    \end{minipage}
    
    \vspace{1em} 
    
    \begin{minipage}{\textwidth}
        \centering
        \includegraphics[width=1\textwidth]{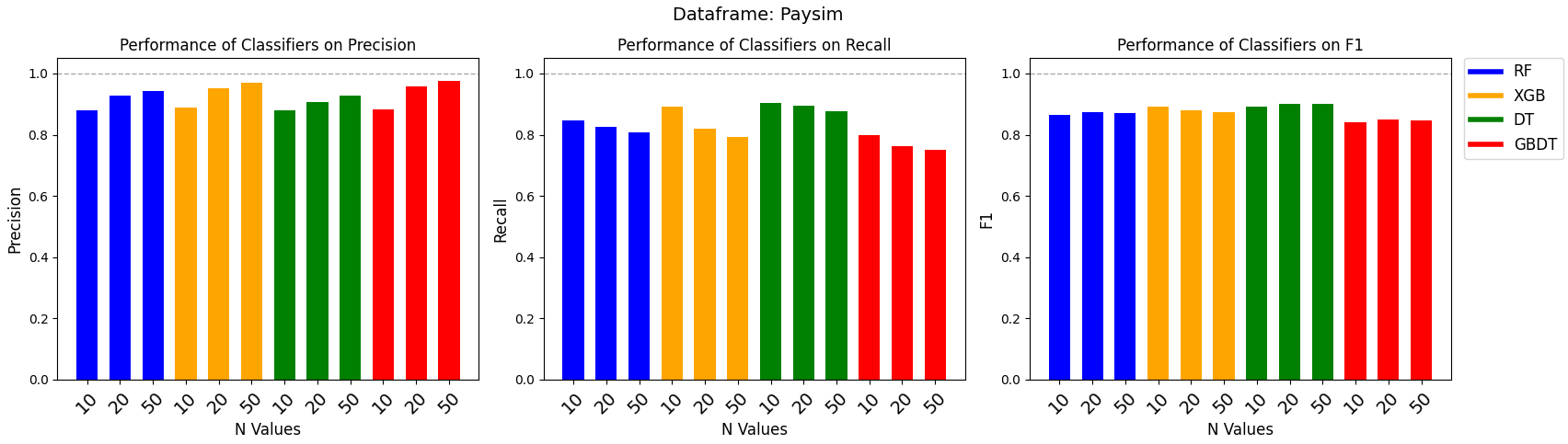} 
    \end{minipage}
    
    \caption{Experiment 3.  SPE and the number of base classifiers  (N = 10, 20, 50) evaluated over Precision, Recall, and F1. The upper row of figures corresponds to the results on Credit Card. The lower row corresponds to the results on PaySim.}
    \label{fig:fig3}
\end{figure}

\begin{figure*}[]
    \centering
    \includegraphics[width=15cm]{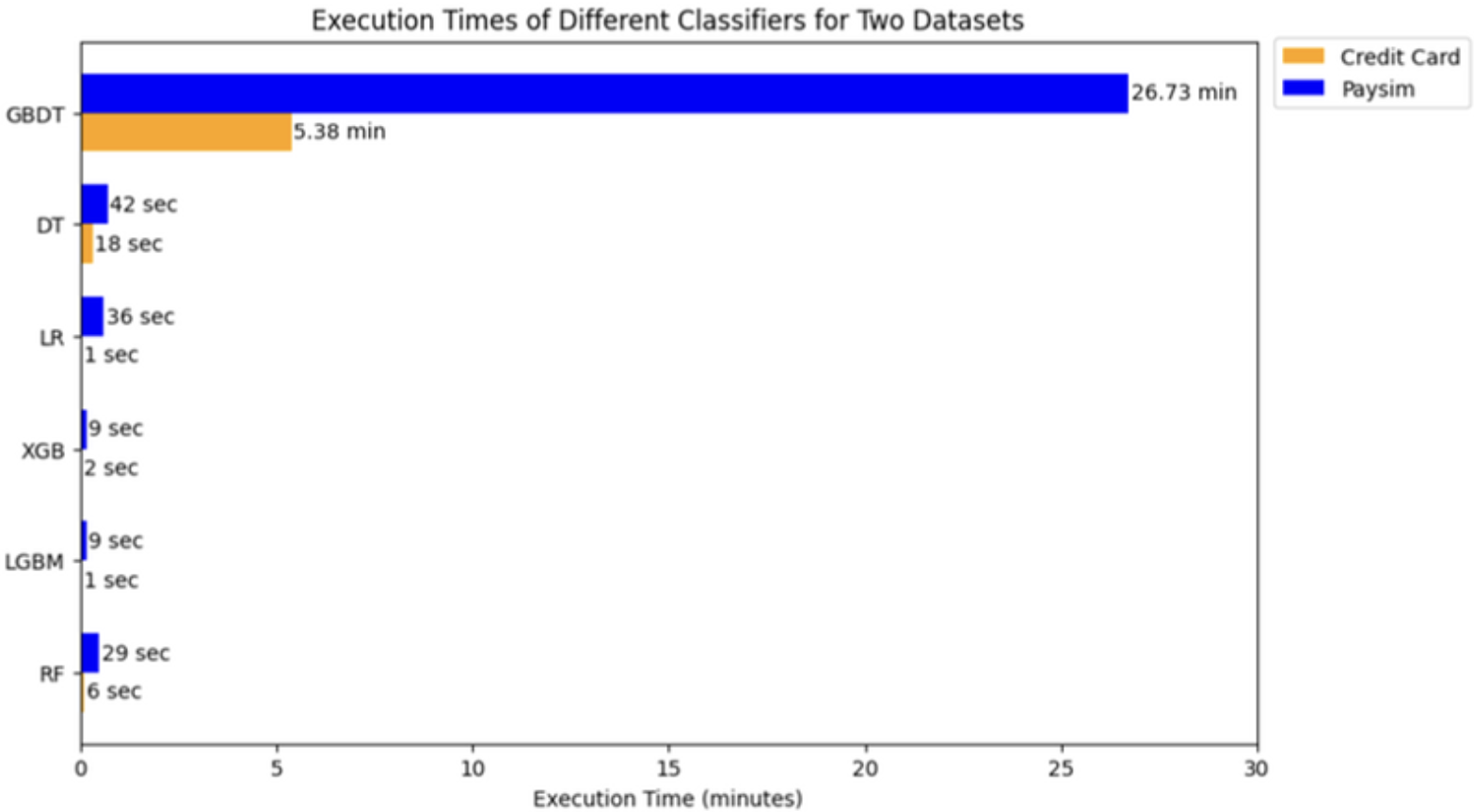}
    \caption{Execution time (minutes:seconds) for a single run including no-sampling, sampling or ensembling when model training. Execution on Google Colab.}.
    \label{f:time}
    
\end{figure*}
Experiment 3 aims at evaluating SPE (ensembling), 
see Figure \ref{fig:fig3}. As in Experiment 2, we evaluate XGB, RF, DT, and GBDT as initial classifiers for SPE with N=10, 20, and 50 classifiers. Interestingly, the more base classifiers, the higher the precision but the lower the recall. On Credit Card dataset, the optimal number of base classifiers is N=20. On PaySim, there are different effects for each classifier. XGB performs best with no more than N=10 base classifiers, while RF, DT, and GBDT do with N=20 base classifiers. Increasing complexity seems unnecessary. 

 On Credit Card, XGB with Over-Sampling ($F1=0.851$, Experiment 2) performs similarly as in ensembling with SPE, N=20 XGB base classifiers ($F1=0.854$, Experiment 3). On PaySim, the best F1 score was obtained by DT No-Sampling ($F1=0.902$, Experiment 1) just like in ensembling with SPE, N=50 DT base classifiers ($F1=0.902$, Experiment 3). These results indicate that it is important to evaluate different classifiers on each problem and dataset. These results align with the No Free Lunch Theorem \cite{wolpert1997no}.

Figure \ref{f:time} presents the execution times. Most of the time, XGB is the fastest, followed by DT, RF, and GBDT, which is the slowest algorithm. Given that under sampling removes samples, it makes execution times the fastest, followed by no sampling, over sampling, and SMOTE. The execution time of SPE increases with the number of base classifiers as expected.

\section{Conclusion}
This study extensively evaluated various machine learning strategies to address the challenge posed by imbalanced datasets, particularly in classification tasks where the minority class is crucial yet underrepresented. We presented three experiments executed on two highly imbalanced and representative datasets containing numerical and categorical features: Credit Card dataset ($283\,726$ samples, imbalance ratio: 598.84:1), 
and PaySim dataset ($6\,362\,620$ samples, imbalance ratio: 773.70:1). 
XGB and RF proved to be robust classifiers and performed well with or without sampling on both datasets in terms of classification accuracy, with XGB being faster than RF. Sampling techniques had positive and negative effects, similarly as reported by previous studies \cite{kim2022, hajek2022fraud, VELARDE2024200354}. Sampling helped improve recall at the expense of deteriorating precision. While in many cases over-sampling the minority class helped improve recognition, under-sampling had a strong deteriorating effect, presumably due to loss of information. Except for GBDT, SMOTE had deteriorating effects, possibly because the generation of synthetic minority samples might introduce noise.  Ensembling with SPE improved recognition in some cases.  Therefore, we recommend evaluating different setups as these might perform differently depending on the dataset characteristics. The findings presented in this study contribute to better understanding and addressing several cybersecurity applications formulated as binary classification problems with imbalanced data, including anomaly detection, fraud detection, intrusion detection, spam detection, or malware detection.


\subsection*{Author contributions}
M.L. experimental design, code, and preparation of tables and figures. G.V. supervision, experimental design, and writing. 

\bibliographystyle{apacite}
\bibliography{references}

\appendix 
\section{Appendix} \label{theappendix}
\begin{table}[H]
    \centering
    \footnotesize 
    \caption{Experiment 1. Performance of individual classifiers. Values close to 1 are better. Tabular representation of the results presented in Figure \ref{fig:fig1}. Best results in bold.}
    \label{tab:3}
    \begin{tabularx}{\textwidth}{@{} l l *{6}{>{\centering\arraybackslash}X} @{}}
        \toprule
        \textbf{Dataset} & \textbf{Metric} & \textbf{RF} & \textbf{LGBM} & \textbf{XGB} & \textbf{LR} & \textbf{DT} & \textbf{GBDT} \\
        \midrule
        Credit Card & Precision & 0.972 & 0.192 & \textbf{0.973} & 0.7 & 0.72 & 0.882\\
                     & Recall & 0.726 & 0.526 & \textbf{0.747} & 0.663 & 0.705 & 0.632 \\
                     & F1 & 0.831 & 0.281 & \textbf{0.845} & 0.681 & 0.713 & 0.736 \\
        \midrule
        Paysim & Precision & \textbf{0.953} & 0.135 & 0.943 & 0.913 & 0.907 & 0.705 \\
                     & Recall & 0.792 & 0.158 & 0.782 & 0.453 & \textbf{0.898} & 0.436 \\
                     & F1 & 0.865 & 0.855 & 0.605 & 0.865 & \textbf{0.902} & 0.539 \\
        \bottomrule
    \end{tabularx}
\end{table}

\begin{table}[H]
    \centering
    \footnotesize 
    \caption{Experiment 2. Effect of sampling the training set, results reported on test set. Values close to 1 are better. Tabular representation of the results presented in Figure \ref{fig:fig2}, Blue, orange, green, and red, marked as (New). Best results in bold.}
    \label{tab:4}
    \begin{tabularx}{\textwidth}{@{} l l l *{7}{>{\raggedright\arraybackslash}X} @{}}
            \toprule
            \textbf{Dataset} & \textbf{Model} & \textbf{Metric} & \textbf{No Sampling} & \textbf{Over-Sampling} & \textbf{SMOTE} & \textbf{Under-Sampling} \\
            \midrule
            Credit Card & RF & Precision & 0.972 & \textbf{0.972} & 0.877 & 0.027 \\
                         &    & Recall & 0.726 & 0.726 & 0.747 & \textbf{0.853} \\
                         &    & F1 & \textbf{0.831} & \textbf{0.831} & 0.807 & 0.052 \\
                         \midrule
                         & XGB & Precision & \textbf{0.973} & 0.937 & 0.864 & 0.026 \\
                         &    & Recall & 0.747 & 0.779 & 0.8 & \textbf{0.895} \\
                         &    & F1 & 0.845 & \textbf{0.851} & 0.831 & 0.051 \\
                         \midrule
                         & DT & Precision & \textbf{0.72} & \textbf{0.72} & 0.442 & 0.02 \\
                         &    & Recall & 0.705 & 0.705 & 0.768 & \textbf{0.853} \\
                         &    & F1 & \textbf{0.713} & \textbf{0.713} & 0.562 & 0.04 \\
                         \midrule
                         & GBDT & Precision & 0.882 & 0.947 & 0.841 & 0.036 \\
                         &    & Recall & 0.632 & 0.747 & 0.779 & \textbf{0.853} \\
                         &    & F1 & 0.736 & \textbf{0.835} & 0.809 & 0.069 \\
            \midrule             
            Paysim & RF & Precision & \textbf{0.953} & 0.927 & 0.646 & 0.093 \\
                         &    & Recall & 0.792 & 0.829 & 0.971 & \textbf{0.999}\\
                         &    & F1 & 0.865 & \textbf{0.875} & 0.776 & 0.17 \\
                         \midrule
                         & XGB & Precision & \textbf{0.943} & 0.612 & 0.458 & 0.118 \\
                         &    & Recall & 0.782 & 0.957 & 0.987 & \textbf{0.998}\\
                         &    & F1 & \textbf{0.855} & 0.747 & 0.625 & 0.211 \\
                         \midrule
                         & DT & Precision & \textbf{0.907} & 0.903 & 0.722 & 0.103 \\
                         &    & Recall & 0.898 & 0.893 & 0.991 & \textbf{0.994} \\
                         &    & F1 & \textbf{0.902} & 0.898 & 0.836 & 0.186 \\
                         \midrule
                         & GBDT & Precision & \textbf{0.705} & 0.468 & 0.391 & 0.134 \\
                         &    & Recall & 0.436 & 0.981 & 0.995 & \textbf{0.999} \\
                         &    & F1 & 0.539 & \textbf{0.634} & 0.562 & 0.236 \\
            \bottomrule
    \end{tabularx}
\end{table}

\begin{table}[H]
    \centering
    \footnotesize 
    \caption{Experiment 3. SPE with N=10, 20, and 50 base classifiers. Tabular representation of the results presented in Figure \ref{fig:fig3}. Best results in bold.}
    \label{tab:6}
    \begin{tabularx}{\textwidth}{@{} l l l *{5}{>{\raggedright\arraybackslash}X} @{}}
        \toprule
        \textbf{Dataset} & \textbf{Method} & \textbf{Base Classifiers} & \textbf{Metric} & \textbf{RF} & \textbf{XGB} & \textbf{DT} & \textbf{GBDT} \\
            \midrule
            Credit Card & SPE & N = 10 & Precision & 0.878 & \textbf{0.925} & 0.525 & 0.9 \\
                         & & & Recall & 0.758 & \textbf{0.779} & 0.768 & 0.758 \\
                         & & & F1 & 0.814 & \textbf{0.846} & 0.624 & 0.823 \\
            \midrule
                                    & & N = 20 & Precision & 0.924 & \textbf{0.961} & 0.833 & 0.947 \\
                                    & &  & Recall & 0.768 & 0.768 & \textbf{0.789} & 0.747 \\
                                    & & & F1 & 0.839 & \textbf{0.854} & 0.811 & 0.835 \\
            \midrule
                                    & & N = 50 & Precision & 0.922 & \textbf{0.959} & 0.947 & 0.959 \\
                                    & & & Recall & \textbf{0.747} & 0.737 & \textbf{0.747} & 0.737 \\
                                    & & & F1 & 0.826 & 0.833 & \textbf{0.835} & 0.833 \\
            \midrule
            Payment Simulation & SPE & N = 10 & Precision & 0.881 & \textbf{0.89} & 0.881 & 0.883 \\
                         & & & Recall & 0.848 & 0.892 & \textbf{0.904} & 0.8 \\
                         & & & F1 & 0.864 & 0.891 & \textbf{0.893} & 0.84 \\
            \midrule
                                    & & N = 20 & Precision & 0.928 & 0.951 & 0.908 & \textbf{0.959} \\
                                    & & & Recall & 0.825 & 0.821 & \textbf{0.894} & 0.764 \\
                                    & & & F1 & 0.873 & 0.881 & \textbf{0.901} & 0.851 \\
            \midrule
                                    & & N = 50 & Precision & 0.943 & 0.969 & 0.928 & \textbf{0.975} \\
                                    & & & Recall   & 0.809 & 0.794 & \textbf{0.877} & 0.751 \\
                                    & & & F1 & 0.871 & 0.873 & \textbf{0.902} & 0.848 \\
            \bottomrule
    \end{tabularx}
\end{table}

\end{document}